\documentclass[sigconf]{acmart}
\setcopyright{none}
\acmConference[WWW '27]{The ACM Web Conference 2027}{May 10--14, 2027}{Dublin, Ireland}
\acmYear{2027}\acmDOI{}\acmISBN{}
\usepackage{amsmath,booktabs,array,tabularx,graphicx,xurl}
\usepackage{tikz}\usetikzlibrary{arrows.meta,positioning,calc,decorations.pathreplacing}
\usepackage{placeins}
\usepackage{stfloats}
\providecommand{\doi}[1]{\url{https://doi.org/#1}}
\newtheorem{claim}{Claim}
\title{Modeling LLM Reasoning Dynamics with Structural Representations}
\author{Ilia Karpov}
\affiliation{\institution{National Research University Higher School of Economics}\city{Moscow}\country{Russia}}
\author{Andrey Atamanyuk}
\affiliation{\institution{National Research University Higher School of Economics}\city{Moscow}\country{Russia}}
\author{Dmitry Ilvovsky}
\affiliation{\institution{National Research University Higher School of Economics}\city{Moscow}\country{Russia}}
\author{Aleksandr Anokhin}
\affiliation{\institution{National Research University Higher School of Economics}\city{Moscow}\country{Russia}}
\author{Alex Elizov}
\affiliation{\institution{Independent researcher}\city{Moscow}\country{Russia}}
\author{Dmitry Zagorulkin}
\affiliation{\institution{National Research University Higher School of Economics}\city{Moscow}\country{Russia}}
\author{Sergei O. Kuznetsov}
\affiliation{\institution{National Research University Higher School of Economics}\city{Moscow}\country{Russia}}
\begin{document}
\begin{abstract}
Language-model reasoning alternates between sustained activities and local operations. We study how this temporal structure predicts continuation and the computation needed to analyze it. T/E/S combines text, local events, and spans of activity. Across reasoning and information-seeking tasks, events and spans improve next-outcome prediction in 70 of 76 evaluation groups for a Transformer and 74 of 76 for Qwen with LoRA. Span duration and preceding events are associated with transitions, and retaining event history across span boundaries improves prediction. We formulate budget-allocation counting and action-conditional continuation queries, prove counting hardness under explicit encodings, and derive conditions for affordable decision certification. In finite-process experiments, bounds that cancel shared continuation mass recover the same optimal actions as full dynamic programming while reducing mean inference CPU time by 77.7\%. On 71 held-out HotpotQA and MuSiQue questions, a compact structural observer improves prediction of whether a fixed inference procedure finishes within a CPU allowance: Brier loss falls from 0.0836 to 0.0756. An annotation audit identifies substantial errors in that observer, separating this computational result from semantic validation of its features.
\end{abstract}
\maketitle
\raggedbottom

\section{Introduction}
A language-model agent can continue reasoning, retrieve evidence, revise a candidate, or answer. The useful action depends on the reasoning already performed. Evaluating the alternatives is itself a computation: a controller may spend more estimating the value of another retrieval round than it saves by avoiding that round. We investigate both the evolving reasoning process and the cost of querying a model of its continuation.

Our representation separates text (T), local events (E), and spans of sustained activity (S). A verification event can occur during derivation without turning the whole interval into verification. Activities can recur, and an event can remain predictive after the activity changes. These distinctions support a duration- and history-dependent approximation of the process. They also suggest a computational question: can the observed prefix predict when analyzing possible continuations will be inexpensive enough to inform an action?

We use \emph{computational availability} operationally. A specified inference procedure must return an estimate or an action certificate within a declared allowance. This concerns a concrete model, algorithm, and input; it does not classify individual prefixes into worst-case complexity classes. Exact probabilities may be difficult to compute even when the preferred action is easy to identify. Conversely, an inexpensive estimate need not improve on an action that a direct text controller already selects.

The project repository is available at \url{https://github.com/karpovilia/reasoning}.

Our contributions are threefold. (1) We characterize duration, local events, and retained history in annotated reasoning and measure their complementary predictive value. (2) We formalize exact counting and budgeted continuation analysis, then specialize decision bounds to shared future states. (3) We evaluate inference-cost prediction and selective analysis on saved paired retrieval continuations, with a separate audit of the structural observer. The counting reductions and concentration arguments apply established theory; the empirical contribution is their connection to observable reasoning and measured inference cost.

Section~\ref{sec:tes} introduces T/E/S and the structural evidence. Section~\ref{sec:theory} defines the computational queries and certificates. Sections~\ref{sec:finite} and~\ref{sec:rag} evaluate finite processes and retrieval decisions. Section~\ref{sec:observation} examines annotation transfer. Proofs and additional comparisons appear in the appendix.

\section{Related work}
ReasonOps segments reasoning using rules and clusters of sentence-opening phrases~\citep{lee2026reasonops}; ReasoningFlow represents discourse relations between steps~\citep{lee2026reasoningflow}. We distinguish local operations from sustained modes, then model elapsed duration and event history within and across modes. Hidden semi-Markov models make duration explicit~\citep{yu2010hsmm}; Hawkes and neural Hawkes processes model event-history effects~\citep{hawkes1971spectra,mei2017neuralhawkes}.

Self-consistency and Tree of Thoughts allocate computation across possible solutions~\citep{wang2023selfconsistency,yao2023tot}. ThoughtTerminator and REFRAIN study stopping generation~\citep{pu2025thoughtterminator,sun2026stop}. Rational metareasoning instead asks which computations are worth performing~\citep{russell1991,hay2012}. Our inference-timing experiment addresses this latter question using the evolving trace, with direct text control as a comparison.

Counting complexity, approximate counting, and sequential uncertainty provide the theoretical tools~\citep{haase2016,gopalan2011,howard2021}. We separate their guarantees from the accuracy of a learned process. Exact inference in a finite representation, or a valid confidence interval within it, does not by itself validate the representation of the underlying language model.

\section{Reasoning structure and prediction}\label{sec:tes}
\paragraph{Representation.}
A \emph{question} is one task; a \emph{trace} is one generated solution. A \emph{segment} is a completed text fragment under a fixed segmentation rule. A \emph{span} is a continuous interval assigned a reasoning mode. Mode and event vocabularies are specified for each benchmark in the annotation instructions. For example, mathematical and scientific tasks use modes such as derivation, exploration, and verification, and events such as backtracking, subgoal completion, and failed attempt. Training and evaluation retain the vocabularies of the corresponding annotation schemes. Several events can occur in one segment. END means natural termination, whereas commitment means adopting an answer and can precede END. A--B--A denotes the order of reasoning modes: for example, derivation $\to$ verification $\to$ derivation again. A and B denote distinct modes; each of the three intervals may span several segments.

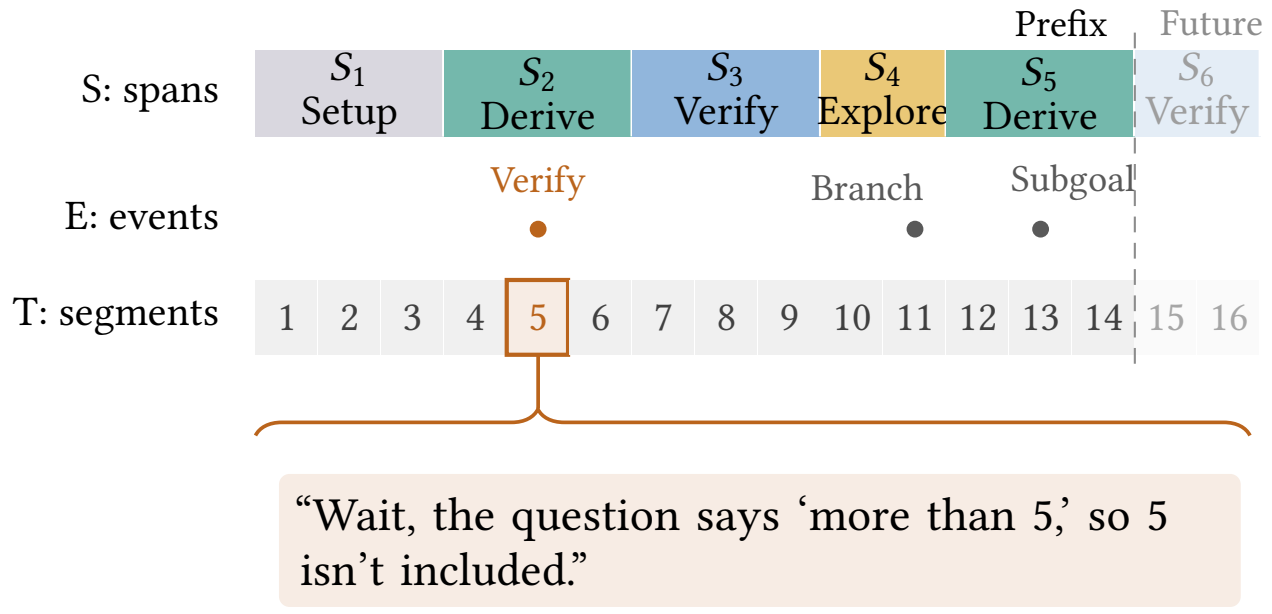
\begin{figure}[t]
\centering\resizebox{\columnwidth}{!}{\usetikzlibrary{decorations.pathreplacing}
\begin{tikzpicture}[x=1cm,y=1cm,font=\fontsize{8}{9.4}\selectfont]
\definecolor{tesSetup}{HTML}{D9D7DF}
\definecolor{tesDerive}{HTML}{74B8AC}
\definecolor{tesVerify}{HTML}{91B6DC}
\definecolor{tesExplore}{HTML}{EAC877}
\definecolor{tesEvent}{HTML}{BA641C}
\path[use as bounding box] (-1.65,-1.65) rectangle (6.76,2.37);
\fill[tesSetup] (0.00,1.50) rectangle (1.26,2.08);
\draw[white,line width=.7pt] (1.26,1.50)--(1.26,2.08);
\node[align=center,text=black] at (0.63,1.79) {$S_1$\\[-1pt]Setup};
\fill[tesDerive] (1.26,1.50) rectangle (2.52,2.08);
\draw[white,line width=.7pt] (2.52,1.50)--(2.52,2.08);
\node[align=center,text=black] at (1.89,1.79) {$S_2$\\[-1pt]Derive};
\fill[tesVerify] (2.52,1.50) rectangle (3.78,2.08);
\draw[white,line width=.7pt] (3.78,1.50)--(3.78,2.08);
\node[align=center,text=black] at (3.15,1.79) {$S_3$\\[-1pt]Verify};
\fill[tesExplore] (3.78,1.50) rectangle (4.62,2.08);
\draw[white,line width=.7pt] (4.62,1.50)--(4.62,2.08);
\node[align=center,text=black] at (4.20,1.79) {$S_4$\\[-1pt]Explore};
\fill[tesDerive] (4.62,1.50) rectangle (5.88,2.08);
\draw[white,line width=.7pt] (5.88,1.50)--(5.88,2.08);
\node[align=center,text=black] at (5.25,1.79) {$S_5$\\[-1pt]Derive};
\fill[tesVerify!25] (5.88,1.50) rectangle (6.72,2.08);
\draw[white,line width=.7pt] (6.72,1.50)--(6.72,2.08);
\node[align=center,text=black!38] at (6.30,1.79) {$S_6$\\[-1pt]Verify};
\node[anchor=east] at (-.13,1.79) {S: spans};
\node[anchor=east] at (-.13,0.97) {E: events};
\node[anchor=east] at (-.13,0.3) {T: segments};
\fill[tesEvent] (1.89,.88) circle (1.6pt);
\node[anchor=south,text=tesEvent,font=\fontsize{7.3}{8.4}\selectfont] at (1.89,.96) {Verify};
\fill[black!65] (5.25,.88) circle (1.6pt);
\node[anchor=south west,text=black!65,font=\fontsize{7.3}{8.4}\selectfont] at (4.94,.96) {Subgoal};
\fill[black!65] (4.41,.88) circle (1.6pt);
\node[anchor=south east,text=black!65,font=\fontsize{7.3}{8.4}\selectfont] at (4.58,.96) {Branch};
\fill[black!6] (0.00,.04) rectangle (0.41,.54);
\node[text=black!75] at (0.21,.29) {1};
\fill[black!6] (0.42,.04) rectangle (0.83,.54);
\node[text=black!75] at (0.63,.29) {2};
\fill[black!6] (0.84,.04) rectangle (1.25,.54);
\node[text=black!75] at (1.05,.29) {3};
\fill[black!6] (1.26,.04) rectangle (1.67,.54);
\node[text=black!75] at (1.47,.29) {4};
\fill[tesEvent!12] (1.68,.04) rectangle (2.09,.54);
\draw[tesEvent,line width=.8pt] (1.68,.04) rectangle (2.09,.54);
\node[text=tesEvent] at (1.89,.29) {5};
\fill[black!6] (2.10,.04) rectangle (2.51,.54);
\node[text=black!75] at (2.31,.29) {6};
\fill[black!6] (2.52,.04) rectangle (2.93,.54);
\node[text=black!75] at (2.73,.29) {7};
\fill[black!6] (2.94,.04) rectangle (3.35,.54);
\node[text=black!75] at (3.15,.29) {8};
\fill[black!6] (3.36,.04) rectangle (3.77,.54);
\node[text=black!75] at (3.57,.29) {9};
\fill[black!6] (3.78,.04) rectangle (4.19,.54);
\node[text=black!75] at (3.99,.29) {10};
\fill[black!6] (4.20,.04) rectangle (4.61,.54);
\node[text=black!75] at (4.41,.29) {11};
\fill[black!6] (4.62,.04) rectangle (5.03,.54);
\node[text=black!75] at (4.83,.29) {12};
\fill[black!6] (5.04,.04) rectangle (5.45,.54);
\node[text=black!75] at (5.25,.29) {13};
\fill[black!6] (5.46,.04) rectangle (5.87,.54);
\node[text=black!75] at (5.67,.29) {14};
\fill[black!2] (5.88,.04) rectangle (6.29,.54);
\node[text=black!35] at (6.09,.29) {15};
\fill[black!2] (6.30,.04) rectangle (6.71,.54);
\node[text=black!35] at (6.51,.29) {16};
\draw[black!45,densely dashed] (5.88,-.02)--(5.88,2.22);
\node[anchor=east,font=\fontsize{7.3}{8.4}\selectfont] at (5.81,2.26) {Prefix};
\node[anchor=west,text=black!45,font=\fontsize{7.3}{8.4}\selectfont] at (5.94,2.26) {Future};
\draw[tesEvent,line width=.7pt] (1.89,.04)--(1.89,-.32);
\draw[tesEvent,line width=.7pt,decorate,decoration={brace,amplitude=5pt,aspect=.284}] (0,-.50)--(6.65,-.50);
\node[anchor=north west,fill=tesEvent!12,rounded corners=2pt,inner sep=4pt,text width=6.15cm,align=left,font=\fontsize{9}{10.5}\selectfont] (excerpt) at (.15,-.75) {``Wait, the question says `more than 5,' so 5 isn't included.''};
\end{tikzpicture}
}
\caption{T/E/S separates a local check from the surrounding derivation. Modes can recur; only the observed prefix is available for prediction.}
\label{fig:tes}
\end{figure}

\paragraph{Data.}
The structural experiments use 56,783 traces from 4555 questions, including 16,497 fully and 997 partially annotated traces. Repeated attempts are available for 4195 questions. Tasks cover MATH/MATH500, BBH, GPQA, HLE, HotpotQA, and MuSiQue~\citep{hendrycks2021math,h4math500,suzgun2022bbh,rein2023gpqa,phan2025hle,yang2018hotpotqa,trivedi2022musique}.

\paragraph{Models.}
Trace generators include Qwen at 4--122B parameters, gpt-oss-20b/120b, DeepSeek-R1 distillations, Gemma, and DeepSeek V4 Flash~\citep{qwen2025,qwenthinking,qwen35large,qwen36,qwen38,oss2025,deepseekr1,gemma4,dsflash0731}. Annotation models assign events and spans; prediction models forecast their continuation; controllers select an action. These roles have different weights and training objectives. The annotation protocol uses Claude Opus 4.8 and DeepSeek Flash/Pro~\citep{claude48,dsflash,dspro}. Appendix~\ref{app:protocols} describes the prediction and observation models.

\paragraph{Duration, events, and returns.}
In 5142 completely annotated traces, transition frequency varies with elapsed mode duration. For BBH it rises from 7.6\% after one segment to 13.5\% after 9--16; for HLE it falls from 54.3\% to 7.2\%. In the broader TRAIN/DEV audit, failed-attempt events precede a next-segment mode change in 36.7\% of cases versus 11.8\% for positions matched on solver, mode, elapsed duration, and trace position. Subgoal completion gives 28.1\% versus 7.4\%. These are conditional associations.

Local events and modes are distinct: 24.7\% of verification events occur outside verification mode. In a separate whole-trace check, 22 verification labels lie inside derivation after excluding run-edge segments. Annotated returns occur in 4021 of the 5142 complete traces (78.2\%). Recurrence is descriptive here; the comparison of samples and annotation protocols is discussed in Limitations.

\paragraph{Complementary predictive roles.}
We predict whether the next segment remains in the current mode, changes mode, or terminates, with destination mode and next-segment events as additional targets. Questions are kept together across repeated solutions, solvers, and annotation versions. On the shared 282-question validation partition, adding E/S to text lowers three-outcome negative log-likelihood (NLL) in \textbf{70 of 76} solver--benchmark--annotation groups for a segment Transformer (Appendix~\ref{app:protocols}) and \textbf{74 of 76} for Qwen with LoRA. NLL is negative natural-log probability, in nats per scored outcome; lower is better. Median relative reductions are 7.0\% and 9.3\%. Both architectures improve in all 32 groups with at least ten evaluation questions.

Table~\ref{tab:ablation} shows the main input ablation. Common position, length, and solver features remain in every row. The best Qwen configuration, including an auxiliary annotation-corruption objective during training, reaches a main multitask score of 0.593, approximately \textbf{0.59}; this score combines transition, destination-mode, and event losses and is not a single-target NLL.

\begin{table}[t]\centering
\caption{Input ablation across corpus solvers: 282 validation questions. Weighted multitask loss; lower is better.}
\label{tab:ablation}
\begin{tabularx}{\linewidth}{@{}Xrr@{}}\toprule
Inputs & Transformer & Qwen\\\midrule
Text & 0.851 & 0.698\\
Text+events & 0.805 & 0.685\\
Text+spans & 0.734 & 0.603\\
Text+events+spans & \textbf{0.719} & \textbf{0.598}\\\bottomrule
\end{tabularx}
\end{table}

Spans provide most of the overall gain; events add predictive information, especially for subsequent events. In the first matched Qwen training, adding events to text and spans lowers next-event binary cross-entropy from 0.281 to 0.249. The 98.75\% paired question-bootstrap interval for the difference is $[-0.042,-0.023]$. The corresponding structural-step NLL changes from 0.502 to 0.499, with interval $[-0.006,0.001]$. Thus the event contribution depends on the predicted target.

\paragraph{An explicit probabilistic approximation.}
We approximate structural continuations using current mode, span age, text state, and two event-memory vectors. Each memory coordinate represents an event type; its update specifies decay and retention across boundaries. This construction permits controlled changes to duration, event generation, and persistence. GRU and Neural Hawkes provide accuracy references on the same prefixes. The explicit model has 11,685 parameters, versus 24,623 and 42,063; GRU has the best next-step NLL and Neural Hawkes the best continuation energy scores (Appendix~\ref{app:models}).

With state-computation parameters fixed and output models refitted, conditioning events on the next mode improves reproduction of event co-occurrence by 9.1\%; retaining accumulated event history improves event counts by mode by 4.6\%. Resetting both memory vectors at boundaries increases next-step NLL from 0.565 to 0.571, with a paired 95\% interval excluding zero. In this experiment, earlier events remain useful after a change of activity.

\section{Counting and choosing computation}\label{sec:theory}
\paragraph{The query.}
For observed prefix $h$, first action $a$, fixed subsequent policy $\pi$, and additional solver budget $B$, define
\begin{equation}
p_M^a(h,B)=\Pr_M(Y=1,C\le B\mid h,\operatorname{do}(a),\pi).
\label{eq:query}
\end{equation}
Here $Y$ denotes a correct terminal answer and $C$ the additional cost. For each query, we fix action $a$, such as retrieving more evidence or answering, and evaluate the success probability under that choice. Utility $Q_M(a)$ can instead be the expectation of a declared bounded payoff. A structural simulator requires an outcome model or scored continuations to connect its events to correctness. Observation and analysis costs are reserved or deducted before fixing the execution budget.

\paragraph{Exact counting.}
Consider $n$ tasks, each with one available additional solution attempt. We choose which attempts to execute within a budget. Attempt $i$ has a binary-encoded integer cost $c_i$ and a rational expected increase $p_i\in[0,1]$ in the probability of solving its task correctly. For budget $B$ and required expected increase $R$ in the number of solved tasks, the number of feasible sets $I\subseteq\{1,\ldots,n\}$ is
\begin{equation}
N(B,R)=\left|\left\{I:\sum_{i\in I}c_i\le B,\quad
\sum_{i\in I}p_i\ge R\right\}\right|.
\label{eq:count}
\end{equation}
\begin{claim}\label{thm:count}
With binary-encoded inputs, counting feasible attempt sets $N(B,R)$ is \#P-complete. Exact computation of the success probability in Equation~\ref{eq:query} is \#P-hard even for finite processes with succinct transitions and polynomially bounded continuation length. Appendix~\ref{app:proofs} specifies the encoding conditions and reductions.
\end{claim}
The proof reduces \#SubsetSum to our query by making the cost and expected gain of each attempt proportional to the same number. The budget and gain constraints then hold together exactly for subsets with the prescribed sum. Appendix~\ref{app:proofs} gives the complete reductions.

For an explicitly enumerated finite model with $E$ edges and horizon $H$, budget dynamic programming costs $O(H(B+1)E)$ arithmetic operations. The bound is pseudo-polynomial in a binary-encoded budget. To select actions, the Bellman recursion computes success probability after each available action, accounting for the remaining budget and steps. It selects the action with the highest success probability. This calculation requires a model of state transitions and terminal outcomes.

\paragraph{Decision certificates.}
For each action, we construct an interval $[L_M(a),U_M(a)]$ containing its expected utility. If one lower bound exceeds every competing upper bound, further refinement cannot change the choice provided the intervals are valid. Values can be estimated by repeatedly simulating a continuation from the same prefix, using fresh independent random numbers in each run. Intervals are constructed to retain their stated coverage probability when computation stops based on the results obtained so far~\citep{howard2021}.

These calculations certify an action under the continuation distribution specified by model $M$. Certification for the real LLM also requires a bound on model error: if the expected utility error for action $a$ is at most $\beta_a$, we expand its interval by $\beta_a$ on each side and check action separation again.

\paragraph{Shared continuations across actions.}
Choosing an action does not always require computing its exact utility. Suppose $Q_0=q(F)/2$ and $Q_1=q(F)/2+1/2$, where $q(F)$ is the probability that a uniformly random assignment satisfies Boolean formula $F$. Computing either $Q$ exactly recovers the number of satisfying assignments, whereas $Q_1-Q_0=1/2$ is known immediately. Selecting action 1 therefore requires no computation of $q(F)$.

In a reasoning model, different first actions can lead to the same future state. When subsequent behavior from that state is identical, the common contribution can be subtracted when comparing actions. Let $g_a$ be the total probability of already evaluated successful continuations, and $\mu_a(x)$ the probability of reaching an unevaluated state $x$ after action $a$. Each unresolved continuation is counted once, at the state where its expansion currently stops. State $x$ includes the remaining budget and steps, event memory, and the rule for subsequent actions. Let $V(x)\in[0,1]$ be the probability of eventual success from $x$. Then $Q_a=g_a+\sum_x\mu_a(x)V(x)$. Put $k=g_a-g_b$ and $d(x)=\mu_a(x)-\mu_b(x)$. We obtain
\begin{equation}
\begin{aligned}
k+\sum_x\min(d(x),0)&\le Q_a-Q_b,\\
Q_a-Q_b&\le k+\sum_x\max(d(x),0).
\end{aligned}
\label{eq:shared}
\end{equation}
The interval width is $\sum_x|d(x)|$. Accounting for shared continuation probability reduces it by $2\sum_x\min(\mu_a(x),\mu_b(x))$ relative to separate action bounds. Cancellation requires matching continuation distributions from the complete state; a common mode name alone does not ensure this.

\section{Reducing action-selection time}\label{sec:finite}
For each action, the model defines the probability of completing the task successfully within the remaining budget. We compare three ways of choosing the action with the highest success probability, testing whether stopping probability computation early preserves the choice while reducing computation time. Finite processes with known transitions provide an exact reference for evaluating early stopping. The experiment covers 32 models and 404 queries from different prefixes. Each query compares three first actions under a fixed subsequent action rule.

The reference method computes each action's success probability in full using dynamic programming. A second method progressively refines separate probability intervals and stops when the best action is determined. A third uses the same stopping criterion but accounts for shared continuations when comparing actions through Equation~\ref{eq:shared}. Their common contribution cancels in the difference, potentially identifying the better action earlier.

Models have 24 or 96 states, horizons of 4, 8, 16, or 32 steps, and budget $\lceil1.5H\rceil$. All methods use exact integer arithmetic. CPU timing covers the full decision calculation: transition processing, storing intermediate results, bound computation, and stopping. Model construction and reference verification are outside the timed calculation. We take the median of five timings per query, then average using fixed instance and configuration weights.

\begin{table}[t]\centering
\caption{Action selection on 404 queries to finite-process models. All methods select the same actions as full computation.}
\label{tab:finite}
\begin{tabularx}{\linewidth}{@{}Xr@{}}\toprule
Method & CPU time, ms $\downarrow$\\\midrule
Full probability computation & 12.94\\
Early selection with separate bounds & 5.62\\
Early selection with shared continuations & \textbf{2.88}\\\bottomrule
\end{tabularx}
\end{table}

The shared-continuation method selects the same actions as full computation in 2.88 rather than 12.94 ms on average, reducing CPU time by 77.7\%. Relative to early selection with separate bounds, the reduction is 48.8\%. The paired instance-bootstrap 95\% interval for the difference from full computation is $[-10.50,-9.63]$ ms. With 24 states and horizon four, accounting for shared continuations is slower than separate bounds because of additional calculations.

\section{Choosing whether to continue retrieval}\label{sec:rag}
\paragraph{Paired actions and models.}
The controller chooses whether to retrieve more evidence or answer from the information already collected. We measure the CPU time needed to compare these actions using a continuation model and test whether that cost can be predicted from the prefix. We use saved HotpotQA and MuSiQue continuations produced by Qwen3.6-35B-A3B-FP8~\citep{qwen36fp8}. After two tool rounds, one branch answers and the other permits up to four further retrieval rounds from the same prefix. Fitting uses 141 questions, development 37, and final evaluation 71 distinct full questions: 36 HotpotQA and 35 MuSiQue, with 194 prefixes and 388 completed branches. Six reserved calibration questions are unused. Whole-question separation and subquestion-family overlap are distinguished in Limitations.

For this computational experiment, a frozen compact MiniLM-based observer produces E/S~\citep{wang2020minilm,minilmcard}. It is a separate model from the main Qwen observer used in earlier prefix-prediction experiments. The process model generates structural continuations and action-conditioned outcome/cost estimates. A separate ridge predictor estimates whether the fixed inference algorithm will obtain a model-relative certificate within an allowance. Text baselines retain hashed text, MiniLM embeddings, counters, and frozen text-based quality/cost forecasts. T+S and T+E+S add observer outputs to those same inputs.

Inference uses up to 1024 paired simulated continuations, horizon 512 structural steps, and a fixed sequence of refinement depths. Utility is answer F1 minus declared input/output-token penalties. We measure three independent random runs per prefix, 582 in evaluation. Availability at 25, 50, 100, or 250 ms means that a certificate was obtained within that measured CPU allowance. A failed computation is missing, not a negative availability label.

\paragraph{Predicting availability.}
Brier loss is the mean squared difference between predicted probability and observed availability. We average over the three runs, all four budgets, prefixes within questions, and then questions equally. All predictors and action rules are fixed before final answer outcomes are opened.

\begin{table}[t]\centering
\caption{Predicting inference availability for Qwen3.6 retrieval: 71 HotpotQA/MuSiQue questions. Lower Brier loss is better.}
\label{tab:availability}
\begin{tabularx}{\linewidth}{@{}Xr@{}}\toprule
Predictor inputs & Brier $\downarrow$\\\midrule
Counters and length & 0.0926\\
Text + counters & 0.0836\\
Text + counters + observed spans & 0.0767\\
Text + counters + observed E/S & \textbf{0.0756}\\\bottomrule
\end{tabularx}
\end{table}

The prespecified E/S-versus-text difference is $-0.0080$, with paired 95\% question-bootstrap interval $[-0.0122,-0.0037]$, a relative reduction of 9.5\%. Most of the gain is already present with spans. This predicts the runtime of a fixed learned computation; the semantic accuracy of its observation layer is evaluated separately in Section~\ref{sec:observation}.

\paragraph{Selecting inference calls.}
At predicted availability at least 0.8, the controller invokes inference; otherwise it uses direct text control. If no certificate is returned, it also uses the direct text action. The four nominal CPU allowances are prediction targets, not hard timeouts: all realized CPU, including overruns, is charged. A matched random rule preserves the number of calls within each benchmark.

\begin{table}[t]\centering
\caption{Qwen3.6 retrieval on 71 HotpotQA/MuSiQue questions: F1, full output tokens, controller CPU. Averages include four budgets.}
\label{tab:routing}
\begin{tabularx}{\linewidth}{@{}Xrrr@{}}\toprule
Rule & F1 $\uparrow$ & Tokens & CPU, ms\\\midrule
Always answer & 0.4339 & 561 & 0.0\\
Always retrieve & 0.5337 & 1410 & 0.0\\
Direct text control & 0.5187 & 1369 & 271.8\\
Inference selected by T & 0.5184 & 1372 & 276.8\\
Inference selected by T/E/S & 0.5192 & 1372 & 278.8\\
Matched random calls & 0.5192 & 1371 & 281.2\\\bottomrule
\end{tabularx}
\end{table}

The E/S selection rule uses 2.34 ms less controller CPU than matched random calls, with 95\% interval $[-4.21,-0.61]$ ms. Their F1 difference is 0.00002, with interval $[-0.00146,0.00175]$. Against direct text control, F1 changes by 0.00058, with interval $[-0.00078,0.00252]$, while CPU increases by 7.05 ms. Mean full input-token counts are 11,225 versus 11,179. Thus predicting inexpensive analysis does not in this experiment produce a better answer-quality/total-cost trade-off than direct text control. CPU measures include observation, feature preparation, prediction, and invoked inference; zero controller CPU for fixed actions does not mean zero generation cost.

\paragraph{Relation to direct computation control.}
Other experiments use structure to choose solver actions directly. On 361 MATH and GPQA questions with two solutions each, E/S reduces Qwen3-4B-Thinking output tokens by 4\% relative to length-and-confidence control and by 0.8\% relative to matched text control, with observed accuracy differences between +0.01 and +0.20 percentage points across the two contrasts and uncertain answer labels. On 65 HotpotQA questions, the structural controller uses 110 tool calls versus 134 for always permitting retrieval, with F1 0.565 versus 0.542. Relative to text control it uses 6\% fewer calls at the same F1, but more output tokens, 1045 versus 984. These experiments establish particular action/cost effects; they use different cohorts and do not evaluate selective probabilistic analysis.

\section{Reliability of structural observation}\label{sec:observation}
To examine the compact observer used in Section~\ref{sec:rag}, DeepSeek V4 Flash labels 72 fixed source segments from 24 questions in two passes with access only to the observed prefix. It sees the question and assistant text only up to the target segment, without detector outputs or answer outcomes. There are 142 valid responses out of 144. Agreement gives known labels for 384 of 432 event cells and 44 mode labels; disagreement is masked. A subsequent source review leaves one additional event cell unresolved.

On the original common mask shared by the compared methods, MiniLM event precision is 0.110, recall 0.421, and F1 0.175, versus F1 0.759 for the earlier source-reviewed labels. An intention filter raises MiniLM F1 only to 0.220. The detector often treats plans to check or search as completed operations. These predictions are present in the process-model inputs, so their errors are relevant to interpretation of the inference-timing experiment. Appendix~\ref{app:annotation} gives the full comparison and the larger single-pass diagnostic. The audit concerns this compact observer; it does not measure the main Qwen detector.

\section{Conclusion}
Reasoning traces exhibit temporal structure that extends beyond isolated steps. Sustained activities differ in duration, local events are associated with transitions, and retained event history improves continuation prediction. T/E/S makes these properties explicit and improves predictive models across tasks and solvers.

The computational analysis separates estimating an exact probability from choosing an action. Shared future-state mass can cancel in a comparison, allowing the same decision with less inference work. On retrieval traces, structural-observer features also predict whether a fixed analysis will finish within a CPU allowance. The action-selection experiment exposes the additional requirement: useful timing must improve a decision enough to repay observation and analysis. Evaluating the main detector against compatible reference annotations and testing revised controllers on fresh question families are the next steps toward that objective.

\section*{Limitations}
The structural prediction results use validation data and several annotation protocols. The 78.2\% return rate and the 33.3\% rate in a separate 90-trace whole-annotation sample describe different task/model compositions and procedures; recurrence frequencies alone do not establish long-range dependence. Machine repeat agreement is not human truth. The compact MiniLM observer has substantial semantic errors, and the main Qwen detector's comparison on the new RAG reference set is not yet complete. Consequently, the inference-timing result supports prediction of the measured computation with its actual features, rather than validated recognition of reasoning events on that domain.

The 71-question retrieval evaluation excludes complete fitting questions, but six of 35 MuSiQue questions share atomic subquestions with fitting, and nine belong to connected fitting families. A post-hoc family bootstrap retains the sign of the Brier difference but does not remove that overlap. Results are specific to a fixed model, inference procedure, and CPU configuration. Exact synthetic certificates do not certify real LLM correctness; the latter requires separate model-error bounds. Retrieval control is evaluated through saved paired branches; common-prefix wall time is unavailable, so absolute end-to-end latency is not reported. The used evaluation questions remain unavailable for retuning or a fresh confirmation.

\section*{Ethical considerations}
We study public reasoning and question-answering benchmarks and saved model-generated continuations. Structural labels describe text, not a model's private internal reasoning or the truth of its statements. Incorrect event detection could cause premature stopping or unnecessary computation, so action quality and cost are reported together. This revision does not introduce deployment on user requests or new claims of human annotation. Generative assistance was used in drafting and translation; empirical values are tied to recorded results.

\FloatBarrier

\FloatBarrier
\appendix
\section{Proofs and inference conditions}\label{app:proofs}
\paragraph{Counting allocations.}
Membership in \#P follows by guessing one inclusion bit per item and checking both rational inequalities in polynomial time. Exact sums of binary-encoded rationals have polynomial bit length. For hardness, use positive-integer \#SubsetSum~\citep{haase2016}. For an instance $a_1,\ldots,a_n,B$, set $A=\sum_i a_i$, $c_i=a_i$, $p_i=a_i/A$, and $R=B/A$. Both bounds hold exactly when $\sum_{i\in I}a_i=B$, giving a bijection between solutions with polynomial encoding length. Trivial empty or zero-total instances can be handled separately. Expected gains are additive over distinct tasks; the proof does not assume additive success probabilities for repeated attempts on the same task.

\paragraph{Continuation probabilities.}
The process state and transition circuit have polynomial encoding length; the horizon and random-bit use are polynomially bounded, and the terminal predicate is computable in polynomial time. For the reduction, the state stores the item index and accumulated integer sum. At each of $n$ steps it chooses a fair bit and adds $a_i$ to both the sum and cost if included. The terminal predicate is sum $\ge B$; feasibility requires cost $\le B$. Thus its success probability is $\#\mathrm{SubsetSum}/2^n$. An exact rational answer recovers the integer count by polynomial-time scaling. More generally, a polynomial fair-bit bound can be padded to a fixed length; counting accepted random strings is in \#P. This does not claim \#P membership for an arbitrary rational-valued function or for processes with unspecified real transitions.

\paragraph{Finite budget dynamics and Bellman recursion.}
Let $V_h(x,b)$ be the probability of success within $h$ further steps and budget $b$. Success is absorbing with value one; failure, negative budget, or exhausted horizon without success has value zero. For a fixed action policy and a finite transition/cost kernel $P$, use
\[
\begin{aligned}
V_h(x,b)&=\sum_{x',c}P(x',c\mid x,a)V_{h-1}(x',b-c),\\
a&=\pi(x,b,h).
\end{aligned}
\]
Optimizing the first and later actions gives the Bellman recursion with $\max_a$ outside the sum. Induction on $h$ proves correctness. Nonnegative integer costs and a finite horizon suffice, including zero-cost transitions. There are $H(B+1)$ copies of each state/edge; rational bit complexity must also be counted. Bounded dependency width offers another route to finite inference~\citep{dechter1999}.

Exact counting hardness does not establish approximation hardness. With independent samples and bounded payoff, additive error $\varepsilon$ requires $O(\varepsilon^{-2}\log(1/\delta))$ samples; generating those samples has its own cost. Relative error for rare success and general succinct inference are different questions. The FPTAS for \#Knapsack concerns its specified counting problem, not every conjunction of budget and success constraints~\citep{gopalan2011}.

\paragraph{Shared-mass certificate.}
Subtract $Q_b$ from $Q_a$. For every state with $d(x)\ge0$, $d(x)V(x)\in[0,d(x)]$; otherwise it lies in $[d(x),0]$. Summing proves Equation~\ref{eq:shared}. Since $|u-v|=u+v-2\min(u,v)$ for nonnegative masses, the width reduction follows. Repeating pairwise comparisons against every competitor certifies the best action. When no strict separation is found, exact expansion can continue to the horizon and apply a fixed tie rule. Model-state equality includes the same continuation policy; merging only equal modes is insufficient.

\paragraph{Sampling and model error.}
For $d$ actions, independent bounded samples in $[0,1]$ within each action, and sample mean $\widehat Q_n(a)$, a conservative radius is
\[
r_n=\sqrt{\frac{\log(\pi^2 d n^2/(3\delta))}{2n}}.
\]
Hoeffding's inequality~\citep{hoeffding1963} and a union bound over $a,n\ge1$ give simultaneous coverage at least $1-\delta$. Adaptive selection of the next action to sample does not change this event. Clipping $\widehat Q_n\pm r_n$ to $[0,1]$ permits optional stopping. Other bounded utility ranges require rescaling; tighter confidence sequences are available~\citep{howard2021}. These are sufficient theoretical conditions, not a claim that measured simulator certificates cover real answer correctness. If $|Q_{\rm real}(a)-Q_M(a)|\le\beta_a$, expand each interval by $\beta_a$; a positive pairwise lower bound must exceed $\beta_a+\beta_b$ to transfer to reality. Learned abstraction errors require the same treatment.

\paragraph{Finite-process implementation.}
The synthetic comparison uses three first actions and fixed subsequent action zero. All transition weights have denominator 32; after $h$ layers masses use denominator $32^h$, permitting exact integer certificates. Shared propagation stores three first-action masses at each common future state. Seeds 128--159 and the full size/horizon grid were fixed before measurement. Confidence intervals resample the 32 instances 5000 times. Exact rational comparisons, zero-budget/horizon cases, and ties were checked outside timing.

\section{Models and experimental protocols}\label{app:protocols}
\paragraph{Predictors and losses.}
The segment Transformer uses frozen 384-dimensional MiniLM embeddings, an 873-to-512 input projection, 12 layers, eight heads, feed-forward width 2048, and a 256-segment window. The separately trained Qwen3-4B-Instruct-2507 predictor uses at most 8192 tokens/256 segments, rank-32 LoRA with $\alpha=64$ and dropout 0.05~\citep{hu2021lora}. Overlapping context is not scored twice. AdamW uses batch 64, weight decay 0.01, clipping at one, 100 warmup updates, then cosine decay. Learning rates are $3\cdot10^{-4}$ for Transformer/projections and $10^{-4}$ for adapters; training uses eight and two passes. The frozen membership comprises 11,058 training traces from 2150 questions and 1498 validation traces from 282; this is the training subset of the corpus, not a competing corpus count.

The reported multitask score is
\[
L=\operatorname{Avg}[I_g\ell_g+(I_g+0.5)I_d\ell_d+0.25I_e\ell_e].
\]
Here $\ell_g$ scores STAY/SWITCH/END, $\ell_d$ the destination mode, and $\ell_e$ known event labels; indicators mask missing targets. Averaging proceeds over annotation namespaces, positions, traces, and questions. Table~\ref{tab:ablation} averages two initializations for Transformer and Qwen T/T+E+S, and one for Qwen T+E/T+S. The best Qwen's auxiliary corruption objective edits up to 5\% of known positions in half the training traces; two auxiliary losses have weight 0.25. The reported 0.593 excludes auxiliary losses. Structural-step comparison uses 239,866 positions from 282 questions; event comparison uses 77,089 positions from 129 questions with known labels.

\paragraph{Annotation and boundaries.}
In prediction experiments, mode duration counts consecutive segments of one mode; unknown labels break the run. An explicit same-mode boundary remains a SWITCH target but does not reset that feature. In the generative model, span age resets at every annotated boundary. The particular process-model comparison uses seven modes, including grounding; the evaluated prefix-detector configuration outputs six modes. END marks natural completion; capped continuations are censored and unknown event labels masked.

The main prefix detector is Qwen3-4B-Instruct-2507~\citep{qweninstruct} with rank-32 LoRA and GRU memory~\citep{cho2014gru}, trained once on 376,153 positions from 500 questions before freezing. Its weights differ from Qwen's continuation predictor. The MiniLM observer in the new timing experiment is a separate compact classifier, not a replacement checkpoint of that detector.

\paragraph{Explicit model comparison.}\label{app:models}
All models in Table~\ref{tab:process} use MATH500, GPQA Diamond, and BBH under a fixed full-annotation partition: 443 fitting questions, 51 internal selection questions, and 82 validation questions. Step NLL is scored on 82 questions; structural energy on 57 and remaining-length/END energy on 59 with known targets. Continuations use 132 saved prefixes and 32 samples per prefix up to 2048 characters. An energy score compares simulated and observed summary vectors using mean distance to the observation minus half the mean pairwise simulation distance. Counts use $x/(1+x)$, lengths the fixed horizon, and structural vectors normalization by $\sqrt{d}$; no validation-fitted scaling is used.

\begin{table}[t]\centering
\caption{Process approximations: MATH500, GPQA Diamond, BBH. NLL in nats/step; two energy scores. Lower is better.}
\label{tab:process}
\begin{tabularx}{\linewidth}{@{}Xrrr@{}}\toprule
Model & NLL & Struct. & Length/END\\\midrule
Semi-Markov & 0.550 & 0.159 & 0.289\\
Explicit memory & 0.546 & 0.159 & 0.315\\
GRU & \textbf{0.529} & 0.160 & 0.272\\
Neural Hawkes & 0.533 & \textbf{0.154} & \textbf{0.269}\\\bottomrule
\end{tabularx}
\end{table}

Mode-conditioned event generation is a mixture of products of Bernoulli probabilities. For event indicator $e\in\{0,1\}$, $\mathrm{Bern}(e;p)=p^e(1-p)^{1-e}$. A shared sampled next mode induces event dependence even when events are conditionally independent within that component. Memory controls retain state parameters and refit outputs in three repetitions, selecting only on internal fitting data. For retained minus reset history, the next-step NLL difference is $-0.0060$, paired 95\% interval $[-0.0109,-0.0016]$, using 10,000 question resamples after averaging repetitions.

\paragraph{Retrieval timing protocol.}
The new evaluation's fitting/\allowbreak development questions yield 398/101 prefixes. Text and structural cost predictors share a ridge penalty chosen by five-fold question validation on fitting data; four output probabilities are clipped and made monotone across allowances. Model utility is F1 minus $0.05\,C_{\rm out}/4096$ and $0.05\,C_{\rm in}/229376$. Refinement depths are 1, 2, 4, \ldots, 512 with nominal model-level error parameter 0.01. Fixed features, process models, cost predictors, and routing threshold precede final evaluation. All comparison rules use the same saved action pairs; no branch is generated in response to a selected rule. The post-hoc family bootstrap groups the final questions into 70 units and yields Brier-difference interval $[-0.0121,-0.0038]$.

The earlier 65-question direct retrieval experiment instead calibrates on 25 HotpotQA and 25 MuSiQue questions. Its text/structure cost weights are 0.2/0.1. Structure minus text gives $-0.108$ tool calls per question, interval $[-0.277,0]$, and +60.7 output tokens, interval $[-7.1,180.2]$, with identical F1. This remains a separate experiment from inference timing.

\section{Reference-annotation audit}\label{app:annotation}
The audit comprises a primary sample with two passes and an extended sample with one pass. The two prefix-based passes are separate requests to the same model using identical annotation instructions. Their common known mask is defined before comparison; disagreements and missing labels are excluded, not negative examples. The primary common-mask comparison contains 373 event cells from 24 questions and 41 mode labels from 19 questions. Event precision, recall, and F1 use pooled confusion counts weighted so that each question contributes total weight one over its known cells.

The earlier reviewed labels achieve precision 0.822, recall 0.704, and F1 0.759. MiniLM gives 0.110, 0.421, and 0.175; with the intention filter, these become 0.149, 0.421, and 0.220. All three comparisons use the same known cells.

A larger single-pass diagnostic has 5642 accepted responses out of 10,345 requested. On 33,542 known event cells across 154 questions, MiniLM precision is 0.150, recall 0.368, and F1 0.213; mode accuracy is 0.410 on 4396 known labels. This is an unfinished, nonrandom slice under one machine pass, so its scores are diagnostic and not interchangeable with the primary two-pass comparison. Together the two stages contain 5784 accepted responses, not 5784 independent traces or questions. Comparison of the main Qwen detector with the fixed primary set remains incomplete; no aggregate Qwen quality score is inferred from these results.

\end{document}